\documentclass[runningheads]{llncs}

\usepackage{times}
\usepackage{url}
\usepackage{latexsym}
\usepackage{booktabs}
\usepackage{pbox}
\usepackage{amsmath}
\usepackage{amsfonts}
\usepackage{adjustbox}
\usepackage{array}
\usepackage[colorlinks,
            linkcolor=black,
            anchorcolor=blue,
            citecolor=blue,
            urlcolor=blue,
            ]{hyperref}
\usepackage[final]{changes}

\usepackage{caption}
\usepackage{subcaption}

\usepackage[misc]{ifsym}

\begin{document}
\setlength{\abovedisplayskip}{4pt}      
\setlength{\belowdisplayskip}{4pt}

\title{Multi-turn Inference Matching Network for Natural Language Inference}

\author{Chunhua Liu\inst{1} \and
Shan Jiang\inst{1}\and
Hainan Yu\inst{1}\and 
Dong Yu \inst{2,1}\Letter }

\authorrunning{C. Liu et al.}
%
\institute{Beijing Language and Culture University \and 
Beijing Advanced Innovation for Language Resources of BLCU
\email{\{chunhualiu596,jiangshan727,ericeryu\}@gmail.com\\
yudong\_blcu@126.com\\}
}

\date{}
\maketitle
\begin{abstract}

Natural Language Inference (NLI) is a fundamental and challenging task in Natural Language Processing (NLP).
Most existing methods only apply one-pass inference process on a mixed matching feature, which is a concatenation of different matching features between a premise and a hypothesis. 
In this paper, we propose a new model called Multi-turn Inference Matching Network (MIMN) to perform multi-turn inference on different matching features. 
In each turn, the model focuses on one particular matching feature instead of the mixed matching feature. To enhance the interaction between different matching features, a memory component is employed to store the history inference information. The inference of each turn is performed on the current matching feature and the memory.
We conduct experiments on three different NLI datasets. The experimental results show that our model outperforms or achieves the state-of-the-art performance on all the three datasets.


\end{abstract}

\section{Introduction}
\label{intro}
Natural Language Inference (NLI) is a crucial subtopic in Natural Language Processing (NLP). 
Most studies treat NLI as a classification problem, aiming at recognizing the relation types of hypothesis-premise sentence pairs, usually including “Entailment”, “Contradiction” and “Neutral”. 

NLI is also called Recognizing Textual Entailment (RTE) \cite{Dagan:2006} in earlier works and a lot of statistical-based \cite{Glickman:2005:PSL:1631862.1631870} and rule-based approaches \cite{MacCartney:EMNLP2008} are proposed to solve the problem. 
In 2015, Bowman released the SNLI corpus \cite{snli:emnlp2015} that provides more than 570K hypothesis-premise sentence pairs. The large-scale data of SNLI allows a Neural Network (NN) based model to perform on the NLI. 
Since then, a variety of NN based models have been proposed, most of which can be divided into two kinds of frameworks. The first one is based on ``Siamense" network \cite{snli:emnlp2015,Mou:2016ACL}. It first applies either Recurrent Neural Network (RNN) or Convolutional Neural Networks (CNN) to generates sentence representations on both premise and hypothesis, and then concatenate them for the final classification.
The second one is called ``matching-aggregation" network \cite{wang-jiang:ICLR2017,zhiguo-wang:ijcai2017}.
It matches two sentences at word level, and then aggregates the matching results to generate a fixed vector for prediction. Matching is implemented by several functions based on element-wise operations \cite{wang-jiang:acl2016,parikh-EtAl:EMNLP2016}. Studies on SNLI show that the second one performs better. 

Though the second framework has made considerable success on the NLI task, there are still some limitations. First, the inference on the mixed matching feature only adopts one-pass process, which means some detailed information would not be retrieved once missing. While the multi-turn inference can overcome this deficiency and make better use of these matching features.
Second, the mixed matching feature only concatenates different matching features as the input for aggregation. It lacks interaction among various matching features. Furthermore, it treats all the matching features equally and cannot assign different importance to different matching features. 

In this paper, we propose the MIMN model to tackle these limitations. 
Our model uses the matching features described in \cite{wang-jiang:ICLR2017,chen-zhu:acl2017}. However, we do not simply concatenate the features but introduce a multi-turn inference mechanism to infer different matching features with a memory component iteratively. The merits of MIMN are as follows: 
\begin{itemize}
\item
MIMN first matches two sentences from various perspectives to generate different matching features and then aggregates these matching features by multi-turn inference mechanism. During the multi-turn inference, each turn focuses on one particular matching feature, which helps the model extract the matching information adequately.
 \item 
MIMN establishes the contact between the current and previous matching features through memory component. The memory component store the inference message of the previous turn. In this way, the inference information flows.

\end{itemize}

We conduct experiments on three NLI datasets: SNLI \cite{snli:emnlp2015}, SCITAIL \cite{Khot:AAAI2018} and MPE \cite{Lai2017NaturalLI}. On the SNLI dataset, our single model achieves 88.3\% in accuracy and our ensemble model achieves 89.3\% in terms of accuracy, which are both comparable with the state-of-the-art results. Furthermore, our MIMN model outperforms all previous works on both SCITAIL and MPE dataset. Especially, the model gains substantial (8.9\%) improvement on MPE dataset which contains multiple premises. This result shows our model is expert in aggregating the information of multiple premises.

\section{Related Work}
Early work on the NLI task mainly uses conventional statistical methods on small-scale datasets \cite{Dagan:2006,Marelli:LREC2014}. 
Recently, the neural models on NLI are based on large-scale datasets and can be categorized into two central frameworks: (i) Siamense-based framework which focuses on building sentence embeddings separately and integrates the two sentence representations to make the final prediction \cite{Mou:2016ACL,Liu:2016CORR,Munkhdalai-Yu:EACL2016,Chen:2017EMNLP,Shen:2017CoRR,Nie:2017EMNLP,tay2017compare:arxiv2018}; 
(ii) ``mat\-ching-aggregation'' framework which uses various matching methods to get the interactive space of two input sentences and then aggregates the matching results to dig for deep information \cite{rocktaschel2016reasoning,pengfei:acl2016,Liu2016ModellingIO,sha-EtAl:2016OLING,parikh-EtAl:EMNLP2016,zhiguo-wang:ijcai2017,McCann:2017,Yu:2017ACL,tay2017compare:arxiv2018,gong2018:iclr2018,ghaeini2018dr:naaclhlt2018}. 

Our model is directly motivated by the approaches proposed by \cite{wang-jiang:acl2016,chen-zhu:acl2017}. 
\cite{wang-jiang:acl2016} introduces the ``matching-aggregation" framework to compare representations between words and then aggregate their matching results for final decision. 

\cite{chen-zhu:acl2017} enhances the comparing approaches by adding element-wise subtraction and element-wise multiplication, which further improve the performance on SNLI. The previous work shows that matching layer is an essential component of this framework and different matching methods can affect the final classification result.

Various attention-based memory neural networks \cite{Weston:ICLR2015} have been explored to solve the NLI problem 
\cite{pengfei:acl2016,cheng-dong-lapata:EMNLP2016,Munkhdalai-Yu:EACL2016}.
\cite{pengfei:acl2016} presents a model of deep fusion LSTMs (DF-LSTMs) (Long Short-Term Memory ) which utilizes a strong interaction between text pairs in a recursive matching memory.  
\cite{cheng-dong-lapata:EMNLP2016} uses a memory network to extend the LSTM architecture.  
\cite{Munkhdalai-Yu:EACL2016} employs a variable sized memory model to enrich the LSTM-based input encoding information. However, all the above models are not specially designed for NLI and they all focus on input sentence encoding. 

Inspired by the previous work, we propose the MIMN model. We iteratively update memory by feeding in different sequence matching features. We are the first to apply memory mechanism to matching component for the NLI task. Our experiment results on several datasets show that our MIMN model is significantly better than the previous models.

\begin{figure}[!t]
\centering
\includegraphics[width=\textwidth]{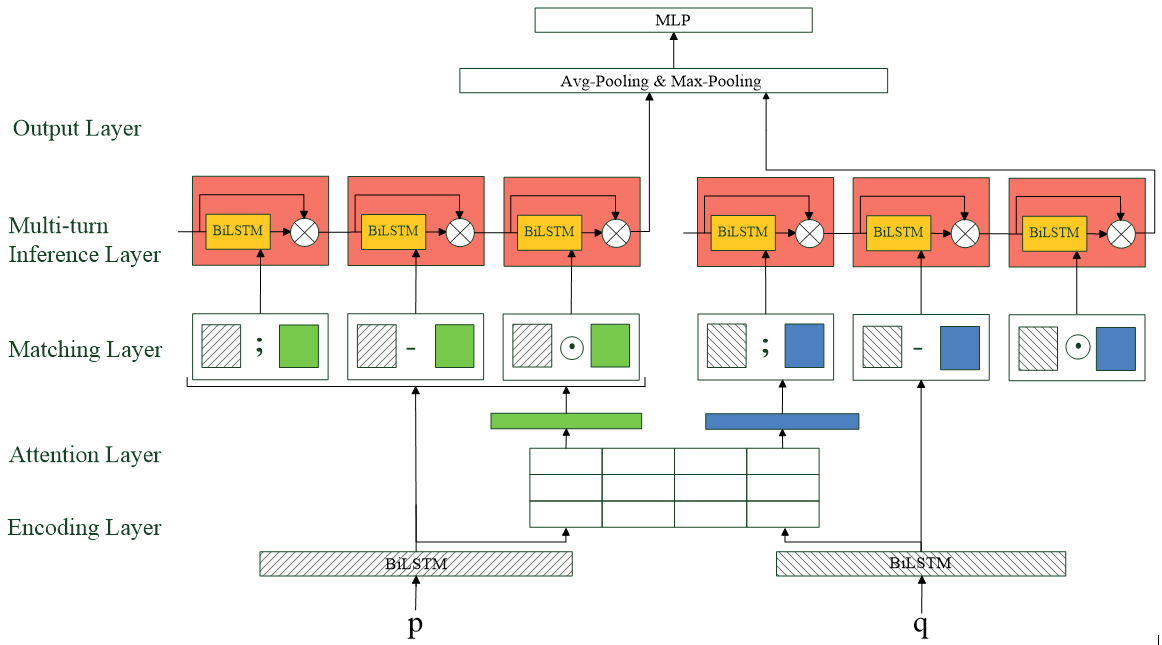}

\caption{\label{fig:model} Architecture  of MIMN  Model.
The matching layer outputs a matching sequence by matching the context vectors with the aligned vectors (green and blue) based on three matching functions.
The multi-turn inference layer generates inference vectors by aggregating the matching sequence over multi-turns.}

\end{figure}

\section{Model}  
\label{sec:model}
In this section, we describe our MIMN model, which consists of the following five major components: encoding layer, attention layer, matching layer, multi-turn inference layer and output layer. Fig.~\ref{fig:model} shows the architecture of our MIMN model.

We represent each example of the NLI task as a triple $(p,q,y)$, where $p = [p_1,p_2, \\ \cdots,p_{l_p}]$ is a given premise, $q = [q_1,q_2,\cdots,q_{l_q}]$ is a given hypothesis, $p_i $ and $q_j \in \mathbb{R}^{r}$ are word embeddings of r-dimension. The true label $y\in\mathcal{Y}$ indicates the logical relationship between the premise $p$ and the hypothesis $q$, where $\mathcal{Y} =\{neutral, entailment, \\  contradiction\}$. Our model aims to compute the conditional probability $Pr(y|p,q)$ and predict the label for examples in testing data set by $y^{*} = argmax_{y\in\mathcal{Y}} Pr(y|p,q) $.

\subsection{Encoding Layer}
In this paper, we utilize a bidirectional LSTM (BiLSTM)  \cite{Hochreiter:1997:LSM:1246443.1246450} as our encoder to transform the word embeddings of premise and hypothesis to context vectors. The premise and the hypothesis share the same weights of BiLSTM.      
\begin{align}
&\bar{p}_i = \text{BiLSTM}_{enc}(p, i) , & i \in [1,2,  \cdots,l_p]  \label{formula:p_i}\\
&\bar{q}_j = \text{BiLSTM}_{enc}(q, j) , & j \in [1,2,\cdots,l_q] \label{formula:q_j}
\end{align}

where the context vectors $\bar{p}_i$ and $\bar{q}_j$ are the concatenation of the forward and backward hidden outputs of BiLSTM respectively. The outputs of the encoding layer are the context vectors $p \in \mathbb{R}^{l_p \times 2d}$ and $q \in \mathbb{R}^{l_q \times 2d}$, where $d$ is the number of hidden units of $\text{BiLSTM}_{enc}$.

\subsection{Attention Layer}  
\label{sect:attention}
On the NLI task, the relevant contexts between the premise and the hypothesis are important clues for final classification. The relevant contexts can be acquired by a soft-attention mechanism \cite{Bahdanau:iclr2015-nmt,Luong:emnlp2015:global-attention}, which has been applied to a bunch of tasks successfully.
The alignments between a premise and a hypothesis are based on a score matrix. There are three most commonly used methods to compute the score matrix: linear combination, bilinear combination, and dot product. For simplicity, we choose dot product in the following computation \cite{parikh-EtAl:EMNLP2016}. 
First, each element in the score matrix is computed based on the context vectors of $\bar{p}_i$ and $\bar{q}_j$ as follows:
\begin{align}
e_{ij} = \bar{p}_i^T \bar{q}_j,
\end{align}
where $\bar{p}_i$ and $\bar{q}_j$ are computed in Equations~(\ref{formula:p_i}) and (\ref{formula:q_j}), and $e_{ij}$ is a scalar which indicates how $\bar{p}_i$ is related to $\bar{q}_j$.

Then, we compute the alignment vectors for each word in the premise and the hypothesis as follows: 
\deleted{
\begin{align}
&\tilde{p}_{i} = \sum_{j=1}^{l_q} \frac{exp(e_{ij})}{\sum{_{t=1}^{l_{q}}exp(e_{it})}} \ \bar{q}_{j},  & \forall i\in [1,...,l_{p}]  \\
&\tilde{q}_{j} = \sum_{i=1}^{l_p} \frac{exp(e_{ij})}{\sum{_{t=1}^{l_{p}}exp(e_{tj})}} \ \bar{p}_{i}, & \forall j\in[1,...,l_{q}]
\end{align}
}
\begin{align}
&\tilde{p}_{i} = \sum_{j=1}^{l_q} \frac{exp(e_{ij})}{\sum{_{t=1}^{l_{q}}exp(e_{it})}} \ \bar{q}_{j},  &
&\tilde{q}_{j} = \sum_{i=1}^{l_p} \frac{exp(e_{ij})}{\sum{_{t=1}^{l_{p}}exp(e_{tj})}} \ \bar{p}_{i}, 
\end{align}
where $\tilde{p_i} \in \mathbb{R}^{2d}$ is the weighted summaries of thehypothesis in terms of each word in the premise. The same operation is applied to $\tilde{q_j} \in \mathbb{R}^{2d}$. 
The outputs of this layer are $\tilde{p_i} \in \mathbb{R}^{l_p \times 2d}$ and $\tilde{q_j} \in \mathbb{R}^{l_q \times 2d}$. 
For the context vectors $\bar{p}$, the relevant contexts in the hypothesis $\bar{q}$ are represented in $\tilde{p}$. The same is applied to $\bar{q}$ and $\tilde{q}$.   

\subsection{Matching Layer}
\label{Matching Layer}
The goal of the matching layer is to match the context vectors $\bar{p}$ and $\bar{q}$ with the corresponding aligned vectors $\tilde{p}$ and $\tilde{q}$ from multi-perspective to generate a matching sequence. 

In this layer, we match each context vector $p_i$ against each aligned vector $\tilde{p}_i$ to capture richer semantic information.
We design three effective matching functions: $f^c$, $f^s$ and $f^m$ to match two vectors \cite{tai-socher-manning:2015:ACL-IJCNLP,wang-jiang:ICLR2017,chen-zhu:acl2017}. 
Each matching function takes the context vector $\bar{p}_i$ ($\bar{q}_j$) and the aligned vector $\tilde{p}_i$ ($\tilde{q}_j$) as inputs, then matches the inputs by an feed-forward network based on a particular matching operation and finally outputs a matching vector. 
The formulas of the three matching functions $f^c$, $f^s$ and $f^m$  are described in formulas~(\ref{formula:f^c})~(\ref{formula:f^s})~(\ref{formula:f^m}). To avoid repetition, we will only describe the application of these functions to $\bar{p}$ and $\tilde{p}$. The readers can infer these equations for $\bar{q}$ and $\tilde{q}$.  
\begin{align}
u^{c}_{p,i}  =  f^c(\bar{p_i},\tilde{p}_i) &= \text{ReLU}(W^c ( [\bar{p_i} \, ; \, \tilde{p}_i]) +b^c) \label{formula:f^c},\\
u^{s}_{p,i}  =  f^s(\bar{p_i}, \tilde{p}_i) &= \text{ReLU}(W^s  ( \bar{p_i} - \tilde{p}_i]) +b^s) \label{formula:f^s},\\
u^{m}_{p,i}  = f^m(\bar{p_i}, \tilde{p}_i)  &= \text{ReLU}(W^m  ( \bar{p_i} \odot \tilde{p}_i]) +b^m) \label{formula:f^m},
\end{align}
where $; \,$, $-$, and $\odot$ represent concatenation, subtraction, and multiplication respectively, $W^c \in \mathbb{R} ^{4d \times d}$, $W^s \in \mathbb{R}^{2d \times d}$ and $W^m \in \mathbb{R}^{2 d \times d}$ are weight parameters to be learned, and $b^c, b^s,  b^m \in \mathbb{R}^d$ are bias parameters to be learned. The outputs of each matching function are $u_{p,i}^c, u_{p,i}^s, u_{p,i}^m \in \mathbb{R}^d $, which represent the matching result from three perspectives respectively. After matching the context vectors $\bar{p}$ and the aligned vectors $\tilde{p}$ by  $f^c$, $f^s$ and $f^m$, we can get three matching features $u_{p}^c = \{u_{p,i}^c\}^{l_p}_1$, $u_{p}^s= \{u_{p,i}^s\}^{l_p}_1$ and $u_{p}^m= \{u_{p,i}^m\}^{l_p}_1$.

\deleted{After matching the context vectors $\bar{p}$ and the aligned vectors $\tilde{p}$ by  $f^c$, $f^s$ and $f^m$, we can get three matching features $u_{p}^c$, $u_{p}^s$ and $u_{p}^m$. where $u_{p}^c = [u_{p,1}^c,u_{p,2}^c, \cdots, u_{p,l_p}^c ]$, $u_{p}^s = [u_{p,1}^s,u_{p,2}^s, \cdots, u_{p,l_p}^{s} ]$, and $u_{p}^m = [u_{p,1}^m,u_{p,2}^m, \cdots, u_{p,l_p}^m ]$.}   

The $u_{p}^c$ can be considered as a joint-feature of combing the context vectors $\bar{p}$ with aligned vectors $\tilde{p}$, which preserves all the information.  
And the $u_{p}^s$ can be seen as a diff-feature of the $\bar{p}$ and $\tilde{p}$, which preserves the different parts and removes the similar parts. 
And the $u_{p}^{m}$ can be regarded as a sim-feature of $p$ and $\bar{p}$, which emphasizes on the similar parts and neglects the different parts between $bar{p}$ and $\tilde{p}$. 
Each feature helps us focus on particular parts between the context vectors and the aligned vectors. 
These matching features are vector representations with low dimension, but containing high-order semantic information. 
To make further use of these matching features, we collect them to generate a matching sequence $u_p$. 
 
\begin{align}
u_p =[u_{p}^1,u_{p}^2,u_{p}^3] = [u_p^c,u_p^s,u_p^m],
\end{align}
where $ u_p^1,u_p^2,u_p^3 \in \mathbb{R}^{l_p \times d}$.  

The output of this layer is the matching sequence $u_p$, which stores three kinds of matching features.
The order of the matching features in $u_p$ is inspired by the attention trajectory of human beings making inference on premise and hypothesis. We process the matching sequence in turn in the multi-turn inference layer. Intuitively, given a premise and a hypothesis, we will first read the original sentences to find the relevant information. Next, it's natural for us to combine all the parts of the original information and the relevant information. Then we move the attention to the different parts. Finally, we pay attention to the similar parts.

\subsection{Multi-turn Inference Layer}
\label{sect:Multi-turn Inference Layer}
In this layer, we aim to acquire inference outputs by aggregating the information in the matching sequence by multi-turn inference mechanism. We regard the inference on the matching sequence as the multi-turn interaction among various matching features.
In each turn, we process one matching feature instead of all the matching features \cite{chen-zhu:acl2017,ghaeini2018dr:naaclhlt2018}. To enhance the information interaction between matching features, a memory component is employed to store the inference information of the previous turns.
Then, the inference of each turn is based on the current matching feature and the memory. Here, we utilize another BiLSTM for the inference.
\begin{align}
&c^{k}_{p,i} = \text{BiLSTM}_{inf}( W_{inf} [u^{k}_{p,i} ; m^{(k-1)}_{p,i}])    \label{formula:BiLSTM_inf},
\end{align}
where $ c^{k}_{p,i} \in \mathbb{R}^{2d}$ is an inference vector in the current turn, $k=[1,2,3]$ is the index current turn, $i=[1,2,3,\cdots,l_p]$, $m^{(k-1)}_{p,i} \in \mathbb{R}^{2d} $ is a memory vector stores the historical inference information, and $W_{inf} \in \mathbb{R}^{3d \times d}$ is used for dimension reduction.                                                           

Then we update the memory by combining the current inference vector $c^{k}_{p,i} $ with the memory vector of last turn $m^{(k-1)i}_p$. An update gate is used to control the ratio of current information and history information adaptively \cite{Wang-nan:2017:ACL}. The initial values of all the memory vectors are all zeros.
\begin{align}
&m^{k}_{p,i}  =   g \odot c^{k}_{p,i} + (1-g)\odot m^{(k-1)}_{p,i}, \label{formula:mem_gate}\\
& \notag g = \sigma( W_g[c^{k}_{p,i} ; m^{(k-1)}_{p,i}] + b_g),
\end{align}
where $W_g \in \mathbb{R}^{4d \times 2d}$ and $b_g \in \mathbb{R}^{2d }$ are parameters to be learned, and $\sigma$ is a sigmoid function to compress the ratio between 0-1. Finally, we use the latest memory matrix $\{m^{3i}_p\}_{1}^{l_p}$ as the inference output of premise $m^{inf}_p$. 
Then we calculate $m^{inf}_q$ in a similar way.
The final outputs of this layer are  $m^{inf}_p$ and $m^{inf}_q$. 
\deleted{
\begin{align}
& m^{inf}_p = \{m^{3}_{p,i}\}_{i=1}^{l_p}  \label{formula:m_inf},
\end{align}

where $m^{inf}_p \in \mathbb{R}^{l_p \times 2d}$ stores the inference results of all matching features.
The final outputs of multi-turn inference layer are  $m^{inf}_p$ and $m^{inf}_q$. The calculation of $m^{inf}_q$ is the same as  $m^{inf}_p$. }

\subsection{Output Layer}
The final relationship judgment depends on the sentence embeddings of premise and hypothesis. We convert $m^{inf}_p$ and $m^{inf}_q$ to sentence embeddings of premise and hypothesis by max pooling and average pooling. Next, we concatenate the two sentence embeddings to a fixed-length output vector. Then we feed the output vector to a multilayer perceptron (MLP) classifier that includes a hidden layer with $tanh$ activation and a softmax layer to get the final prediction. The model is trained end-to-end. We employ multi-class cross-entropy as the cost function when training the model.

\section{Experiment}
\label{sec:exp}
\subsection{Data}
To verify the effectiveness of our model, we conduct experiments on three NLI datasets. The basic information about the three datasets is shown in Table~\ref{table:datasets}. 

The large SNLI \cite{snli:emnlp2015} corpus is served as a major benchmark for the NLI task.
The MPE corpus \cite{Lai2017NaturalLI} is a newly released textual entailment dataset. 
Each pair in MPE consists of four premises, one hypothesis, and one label, which is different from the standard NLI datasets. Entailment relationship holds if the hypothesis comes from the same image as the four premises.
The SCITAIL \cite{Khot:AAAI2018} is a dataset about science question answering. The premises are created from relevant web sentences, while hypotheses are created from science questions and the corresponding answer candidates.
\vspace{-0.4cm}
\begin{table}[!h]
\begin{center}
\begin{adjustbox}{max width=\textwidth}
\begin{tabular}{|p{18mm}<{\centering}|p{20mm}<{\centering}|p{18mm}<{\centering}|p{18mm}<{\centering}|p{18mm}<{\centering}|p{18mm}<{\centering}|}
\hline
Dataset& Sentence Pairs  & Train & Valid & Test &Labels  \\
\hline   
SNLI & 570k & 549,367 & 9,842 & 9,824 & N, E, C \\
\hline
MPE & 10k & 8,000 & 1,000 & 1,000 & N, E, C \\
\hline
SCITAIL & 24k & 23,596 & 1,304 & 2,126 & N, E \\
\hline
\end{tabular}
\end{adjustbox}
\vspace{1em}
\setlength{\belowcaptionskip}{-1.cm}
\caption{ \label{table:datasets} Basic information about the three NLI datasets. Sentence Pairs is the total examples of each dataset. 
\deleted{Train, Valid, and Test are the dataset partition for training, validation and testing.} N, E, and C indicate Neutral, Entailment, and Contradiction, respectively.}  
\label{nli_datasets}
\end{center}
\end{table}

\subsection{ Models for Comparison}

We compare our model with ``matching-aggregation'' related and attention-based memory related models. In addition, to verify the effectiveness of these major components in our model, we design the following model variations for comparison:
\deleted{ESIM is considered as a typical model of ``matching-aggregation'', so we choose ESIM as the principal comparison object. We choose the LSTMN model with deep attention fusion as a complement comparison, which is a memory related model. Besides above models, following variants of our model are designed for comparing:} 

\begin{itemize}

\item {\bf ESIM } We choose the ESIM model as our baseline. It mixes all the matching feature together in the matching layer and then infers the matching result in a single-turn with a BiLSTM.  

\item {\bf 600D MIMN}:
This is our main model described in section~\ref{sec:model}.

\item {\bf 600D MIMN-memory}: This model removes the memory component. 
The motivation of this experiment is to verify whether the multiple turns inference can acquire more sufficient information than one-pass inference. In this model, we process one matching feature in one iteration. 
The three matching features are encoded by $\text{BiLSTM}_{inf}$ in multi-turns iteratively without previous memory information. 
The output of each iteration is concatenated to be the final output of the multi-turn inference layer:  
\deleted{Then the Equation~(\ref{formula:BiLSTM_inf}) and~(\ref{formula:m_inf}) are changed into Equation~(\ref{formula:BiLSTM_inf_no_mem}) and~(\ref{formula:m_inf_no_mem}) respectively and the Equation~(\ref{formula:mem_gate}) is removed.}
\begin{align}
&c^{k}_{p,i} = \text{BiLSTM}_{inf}(W_{inf}[u^{k}_{p,i}]), \label{formula:BiLSTM_inf_no_mem} \\
&m^{inf}_p =[\{c^1_{p,i}\}_{1}^{l_p};\{c^2_{p,i}\}_{1}^{l_p};\{c^3_{p,i}\}_{1}^{l_p}]. \label{formula:m_inf_no_mem}
\end{align}

\item {\bf 600D MIMN-gate+ReLU }:  This model replaces the update gate in the memory component with a ReLU layer. The motivation of this model is to verify the effectiveness of update gate for combining current inference result and previous memory. Then the Equation~(\ref{formula:mem_gate}) is changed into Equation~(\ref{formula:mem_relu}).
\deleted{$m^{inf}_p$ stays the same as Equations~(\ref{formula:m_inf}).}

\begin{align}
&m^{k}_{p,i} = \text{ReLU}(W_m [c^{k}_{p,i} ; m^{(k-1)}_{p,i}] ). \label{formula:mem_relu}
\end{align}

\end{itemize}

\subsection{Experimental Settings}   
We implement our model with Tensorflow \cite{Abadi:2016}. We initialize the word embeddings by the pre-trained embeddings of 300D GloVe 840B vectors \cite{Pennington:emnlp2014}. 
The word embeddings of the out-of-vocabulary words are randomly initialized. The hidden units of $\text{BiLSTM}_{enc}$ and $\text{BiLSTM}_{inf}$ are 300 dimensions.  All weights are constrained by L2 regularization with the weight decay coefficient of 0.0003. We also apply dropout \cite{srivastava:JMLR2014} to all the layers with a dropout rate of 0.2. Batch size is set to 32. 
The model is optimized with Adam \cite{Kingma:ICLR2014} with an initial learning rate of 0.0005, the first momentum of 0.9 and the second of 0.999.  
The word embeddings are fixed during all the training time. We use early-stopping (patience=10) based on the validation set accuracy. We use three turns on all the datasets. 
The evaluation metric is the classification accuracy.
To help duplicate our results, we will release our source code at \url{https://github.com/blcunlp/RTE/tree/master/MIMN}.

\subsection{Experiments on SNLI}
Experimental results of the current state-of-the-art models and three variants of our model are listed in Table~\ref{res:snli}. The first group of models (1)-(3) are the attention-based memory models on the NLI task. 
\cite{pengfei:acl2016} uses external memory to increase the capacity of LSTMs. \cite{Munkhdalai-Yu:EACL2016} utilizes an encoding memory matrix to maintain the input information.  \cite{cheng-dong-lapata:EMNLP2016} extends the LSTM architecture with a memory network to enhance the interaction between the current input and all previous inputs. 

The next group of models (4)-(12) belong to the ``matching-aggregation'' framework with bidirectional inter-attention. 
Decomposable attention \cite{parikh-EtAl:EMNLP2016} first applies the ``matching-aggregation'' on SNLI dataset explicitly. 
\cite{wang-jiang:ICLR2017} enriches the framework with several comparison functions. 
BiMPM \cite{zhiguo-wang:ijcai2017} employs a multi-perspective matching function to match the two sentences.
\deleted{BiMPM \cite{zhiguo-wang:ijcai2017} does not only exploit a multi-perspective matching function but also allows the two sentences to match from multi-granularity.} 
ESIM \cite{chen-zhu:acl2017} further sublimates the framework by enhancing the matching tuples with element-wise subtraction and element-wise multiplication. ESIM achieves 88.0\% in accuracy on the SNLI test set, which exceeds the human performance (87.7\%) for the first time. 
\cite{tay2017compare:arxiv2018} and \cite{Glickman:2005:PSL:1631862.1631870} both further improve the performance by taking the ESIM model as a baseline model. 
The studies related to ``matching-aggregation'' but without bidirectional interaction are not listed \cite{rocktaschel2016reasoning,wang-jiang:acl2016}.

Motivated by the attention-based memory models and the bidirectional inter-attention models, we propose the MIMN model. The last group of models (13)-(16) are models described in this paper. 
Our single MIMN model obtains an accuracy of 88.3\%  on SNLI test set, which is comparable with the current state-of-the-art single models. 
The single MIMN model improves 0.3\%  on the test set compared with ESIM, which shows that multi-turn inference based on the matching features and memory achieves better performance. 
From model (14), we also observe that memory is generally beneficial, and the accuracy drops 0.8\% when the memory is removed. 
This finding proves that the interaction between matching features is significantly important for the final classification. 
To explore the way of updating memory, we replace the update gate in MIMN with a ReLU layer to update the memory, which drops 0.1\%. 
\begin{table}[!tbp]
\renewcommand{\arraystretch}{0.9}
\begin{center}
\begin{adjustbox}{max width=\textwidth}
\begin{tabular}{lccc}
\toprule
Model (memory related) & Parameters & Train(\% acc) & Test(\% acc) \\
\midrule
(1) 100D DF-LSTM \cite{pengfei:acl2016}	&320k	&85.2	&84.6\\
(2) 300D MMA-NSE with attention \cite{Munkhdalai-Yu:EACL2016}	&3.2m	&86.9	&85.4\\
(3) 450D LSTMN with deep attention fusion \cite{cheng-dong-lapata:EMNLP2016}	&3.4m	&88.5	&86.3\\
\midrule 
Model (bidirectional inter-attention) & Parameters & Train(\% acc) & Test(\% acc) \\
\midrule  
(4) 200D decomposable attention \cite{parikh-EtAl:EMNLP2016}	&380k	&89.5	&86.3\\
(5) ``compare-aggregate'' \cite{wang-jiang:ICLR2017} & - & 89.4 & 86.8\\
(6) BiMPM \cite{zhiguo-wang:ijcai2017}	&1.6m	&90.9	&87.5\\
(7) 600D ESIM \cite{chen-zhu:acl2017} &4.3M & 92.6 &88.0 \\
(8) 300D CAFE \cite{tay2017compare:arxiv2018}	&4.7m	&89.8	& {\bf 88.5}\\
(9) 450D DR-BiLSTM \cite{ghaeini2018dr:naaclhlt2018}	&7.5m	&94.1	&{\bf 88.5}\\
(10) BiMPM (Ensemble)	\cite{zhiguo-wang:ijcai2017} &6.4m	&93.2	&88.8\\
(11) 450D DR-BiLSTM (Ensemble) \cite{ghaeini2018dr:naaclhlt2018}&45m	&94.8	&89.3\\
(12) 300D CAFE (Ensemble) \cite{tay2017compare:arxiv2018}&17.5m	&92.5	&89.3\\
\midrule  
Human Performance (Estimated) & - & - & 87.7 \\
\midrule  
Model (this paper) & Parameters & Train(\%acc) & Test(\%acc) \\
\midrule  
(13) 600D MIMN & 5.3m &92.2 & {\bf 88.3} \\
(14) 600D MIMN-memory & 5.8m  &87.5 & 87.5 \\ 
(15) 600D MIMN-gate+ReLU & 5.3m  &90.7 & 88.2\\
(16) 600D MIMN ({\bf Ensemble }) & - &92.5 & {\bf 89.3} \\
\bottomrule  
\end{tabular}
\end{adjustbox}
\vspace{1em}
\setlength{\belowcaptionskip}{-0.3cm}
\caption{\label{res:snli} Performance on SNLI}
\end{center}
\end{table}

To further improve the performance\deleted{ on SNLI dataset}, an ensemble model MIMN is built for comparison. We design the ensemble model by simply averaging the probability distributions \cite{zhiguo-wang:ijcai2017} of four MIMN models. Each of the models has the same architecture but initialized by different seeds.
Our ensemble model achieves the state-of-the-art performance by obtains an accuracy of 89.3\% on SNLI test set.

\subsection{Experiments on MPE}

The MPE dataset is a brand-new dataset for NLI with four premises, one hypothesis, and one label. In order to maintain the same data format as other textual entailment datasets (one premise, one hypothesis, and one label), we concatenate the four premises as one premise. 

Table~\ref{res:mpe} shows the results of our models along with the published models on this dataset. 
LSTM is a conditional LSTM model used in \cite{rocktaschel2016reasoning}. 
WbW-Attention aligns each word in the hypothesis with the premise. 
The state-of-the-art model on MPE dataset is SE model proposed by \cite{Lai2017NaturalLI}, which makes four independent predictions for each sentence pairs, and the final prediction is the summation of four predictions. Compared with SE, our MIMN model obtains a dramatic improvement (9.7\%) on MPE dataset by achieving 66.0\% in accuracy.

To compare with the bidirectional inter-attention model, we re-implement the ESIM, which obtains 59.0\% in accuracy. We observe that MIMN-memory model achieves 61.6\% in accuracy. This finding implies that inferring the matching features by multi-turns works better than single turn. Compared with the ESIM, our MIMN model increases 7.0\% in accuracy. 
We further find that the performance of MIMN achieves 77.9\% and 73.1\% in accuracy of entailment and contradiction respectively, outperforming all previous models.  From the accuracy distributions on N, E, and C in Table~\ref{res:mpe}, we can see that the MIMN model is good at dealing with entailment and contradiction while achieves only average performance on neural.  

Consequently, the experiment results show that our MIMN model achieves a new state-of-the-art performance on MPE test set. 
\deleted{Besides, our MIMN-memory model and MIMN-gate+ReLU model both achieve better performance than previous models.}  
All of our models perform well on the entailment label, which reveals that our models can aggregate information from multiple sentences for entailment judgment. 

\begin{table}[!tbp]
\begin{adjustbox}{max width=\textwidth}
\parbox[]{.52 \linewidth}{\hrule height 0pt width 0pt 
\begin{center}
\renewcommand{\arraystretch}{1.08}
\begin{tabular}{lcccc}
\toprule
Models & Test(\%acc)  & N & E & C\\
\midrule
$\text{LSTM}^\#$    & 53.5 & \textbf{39.2} & 63.1 &53.5 \\
$\text{WbW-Attention}^\#$ & 53.9 & 30.2 & 61.3 & 66.5 \\
$\text{SE}^\#$ & \textbf{56.3} & 30.6 & 48.3 & 71.2 \\
\midrule
ESIM (our\_imp) & 59.0 & 34.1 &68.3 &65.1	\\
\midrule
MIMN & \textbf{66.0}& 35.3 &  \textbf{77.9} & \textbf{73.1}  \\
MIMN-memory & 61.6 & 28.4 &72.7 &70.8\\
MIMN-gate+ReLU & 64.8 & 37.5 & 77.9 & 69.1\\
\bottomrule
\end{tabular}
\vspace{1em}
\caption{\label{res:mpe} Performance on MPE. Models with $^\#$ are reported from \cite{Lai2017NaturalLI}.}
\end{center}
}
\hfill
\parbox[]{.55\linewidth}{\hrule height 0pt width 0pt
\begin{center}
\begin{tabular}{lcc}
\toprule
Models &Valid(\%acc) & Test(\%acc) \\
\midrule  
$\text{Majority class}^\star $ & 63.3 & 60.3\\
$\text{decomposable attention}^\star$ & 75.4 & 72.3\\
$\text{ESIM}^\star$  & 70.5 & 70.6\\
$\text{Ngram}^\star$  & 65.0 & 70.6\\
$\text{DGEM}^\star$  & 79.6 & 77.3 \\
CAFE \cite{tay2017compare:arxiv2018}& - & 83.3 \\
\midrule
MIMN & 84.7& \textbf{84.0}\\
MIMN-memory & 81.3 & 82.2  \\
MIMN-gate+ReLU & 83.4  & 83.5  \\
\bottomrule  
\end{tabular}
\vspace{1em}
\caption{\label{res:scitail} Performance on SCITAIL. Models with $^\star$ are reported from \cite{Khot:AAAI2018}.}
\end{center}
}
\end{adjustbox}
\end{table}

\subsection{Experiments on SCITAIL}
In this section, we study the effectiveness of our model on the SCITAIL dataset.
Table~\ref{res:scitail} presents the results of our models and the previous models on this dataset. Apart from the results reported in the original paper \cite{Khot:AAAI2018}: Majority class, ngram, decomposable attention, ESIM and DGEM, we compare further with the current state-of-the-art model CAFE \cite{tay2017compare:arxiv2018}.   

We can see that the MIMN model achieves 84.0\% in accuracy on SCITAIL test set, which outperforms the CAFE by a margin of 0.5\%. Moreover, the MIMN-gate+ReLU model exceeds the CAFE slightly. 
The MIMN model increases 13.3\% in test accuracy compared with the ESIM, which again proves that multi-turn inference is better than one-pass inference.

\section{Conclusion}
In this paper, we propose the MIMN model for NLI task. Our model introduces a multi-turns inference mechanism to process multi-perspective matching features. Furthermore, the model employs the memory mechanism to carry proceeding inference information. In each turn, the inference is based on the current matching feature and previous memory. 
Experimental results on SNLI dataset show that the MIMN model is on par with the state-of-the-art models. 
Moreover, our model achieves new state-of-the-art results on the MPE and the SCITAL datasets. 
Experimental results prove that the MIMN model can extract important information from multiple premises for the final judgment.
And the model is good at handling the relationships of entailment and contradiction.    

\section*{Acknowledgements}

This work is funded by Beijing Advanced Innovation for Language Resources of BLCU, the Fundamental Research Funds for the Central Universities in BLCU (No.17PT05) and Graduate Innovation Fund of BLCU (No.18YCX010).

\bibliographystyle{splncs04}
\bibliography{splncs04}


\end{document}